# Nonlinear Intensity Underwater Sonar Image Matching Method Based on Phase Information and Deep Convolution Features

Xiaoteng Zhou, Changli Yu, Xin Yuan, Haijun Feng, and Yang Xu

*Abstract*—In the field of deep-sea exploration, sonar is presently the only efficient long-distance sensing device. The complicated underwater environment, such as noise interference, low target intensity or background dynamics, has brought many negative effects on sonar imaging. Among them, the problem of nonlinear intensity is extremely prevalent. It is also known as the anisotropy of acoustic sensor imaging, that is, when autonomous underwater vehicles (AUVs) carry sonar to detect the same target from different angles, the intensity variation between image pairs is sometimes very large, which makes the traditional matching algorithm almost ineffective. However, image matching is the basis of comprehensive tasks such as navigation, positioning, and mapping. Therefore, it is very valuable to obtain robust and accurate matching results. This paper proposes a combined matching method based on phase information and deep convolution features. It has two outstanding advantages: one is that the deep convolution features could be used to measure the similarity of the local and global positions of the sonar image; the other is that local feature matching could be performed at the key target position of the sonar image. This method does not need complex manual designs, and completes the matching task of nonlinear intensity sonar images in a close end-to-end manner. Feature matching experiments are carried out on the deep-sea sonar images captured by AUVs, and the results show that our proposal has preeminent matching accuracy and robustness.

## I. INTRODUCTION

IN recent years, in order to meet the sustainable development of human and industry, deep-sea exploration activities have become more and more frequent. In the process of underwater detection, due to the absorption and scattering of natural light by the medium, the effective distance of optical sensors is seriously limited. In the process of deep-water detection using AUVs and other apparatus, acoustic imaging devices, such as side scan sonar (SSS), are widely used sensors, principally because acoustic waves are the most effective way of underwater long-distance detection at present. Detection activities may include independent measurement using multiple platforms and multiple sonar systems, which may have different operation frequencies, detection viewpoints and operation dates. These differences will lead to the nonlinear intensity difference characteristics of sonar images, especially the dependence on detection viewpoints. In short, for the same target, sonar detects from different viewpoints, there are evident differences in the images obtained. Some scholars also call it anisotropy of acoustic imaging, which is used to represent the sensitivity of intensity to viewpoint [1-4]. The non-linear intensity difference of sonar images described above seriously hinders the development of sonar image matching technology. Sonar image matching technology plays a key role in the construction of submarine maps [5], autonomous navigation of underwater vehicles [1] and autonomous docking of underwater mobile equipment [6]. In order to completely excavate the information within the sonar image, and serve it for the comprehensive AUV underwater detection task, effectively solve the problem of the nonlinear intensity of the sonar image to further enhance the matching accuracy of the sonar image, which has long-term significance.

The rest of this letter is organized as follows. Section II introduces the related work of matching underwater sonar image. Section III details our proposed methodology. Section IV states the details of our experiments and tests. The evaluation is given in Section V. The conclusions are drawn in Section VI.

## II. RELATED WORK

Due to the dependence on viewpoints, the matching of sonar images has always been a difficult problem to solve. The mainstream research ideas are chiefly divided into three types: (i) Based on the classic matching algorithms, such as the scale-invariant feature transform (SIFT) [7], these algorithms are usually developed for natural optical images. (ii) Manually design a matching algorithm for sonar images, such as the texture statistics and shadow statistics. (iii) Introduce a convolutional neural network (CNN) method to match in a data-driven manner. In [8], the author compares the performance of

This research was supported by the Chinese Shandong Provincial Key Research and Development Plan, under Grant No. 2019GHZ011 and NO. 2021CXGC010702. *(Corresponding author: Changli Yu and Xin Yuan.)*

Xiaoteng Zhou, Changli Yu, Xin Yuan, Haijun Feng are with the School of Ocean Engineering, Harbin Institute of Technology, Weihai 264209, China (e-mail: zhouxiaoteng@stu.hit.edu.cn; yuchangli@hitwh.edu.cn; xin.yuan@upm.es; fenghaijun77@gmail.com).

Yang Xu is with the School of Automotive Engineering, Harbin Institute of Technology, Weihai 264209, China (e-mail: xuyangxy7@gmail.com).

classic matching algorithms on SSS images, and gives complete experimental data. The matching algorithms include the SIFT, SURF [9], ORB [10], etc. The results show that when there is no large nonlinear intensity difference in the sonar image, SIFT and SURF work best. [1] compared the matching performance on SSS images based on mutual information maximization, logarithmic-polarity cross-correlation, and SIFT, and evaluates it through a series of indicators such as execution time and matching accuracy. The result shows that SIFT has better performance. In [6,11], the authors manually design a matching method through prior knowledge and an expert system, which can basically meet the needs of specific underwater matching tasks. In recent years, the CNN has made great achievements in the field of image processing, and application scenarios such as: target recognition, detection and tracking [12]. In the field of underwater detection research, some researchers try to use CNN to solve the matching problem of sonar images, in [3], the author proposed to use CNN to establish a specific similarity evaluation model to solve the matching problem of forward-looking sonar (FLS) images. In this research, a certain number of FLS image datasets are collected for training and testing. The results show that CNN matches. The performance is better than the classic SIFT, SURF and other algorithms. This research is a successful attempt to introduce the CNN network into the sonar image matching task. The author in [13] proposed to use the CNN network to establish a similarity evaluation model to solve the SSS image matching problem, and ultimately to serve the AUV autonomous navigation. The research tried several SSS image matching tasks from different detection viewpoints, and achieved high accuracy.

In a real underwater operation scene, the sensor carrying platform such as AUV will inevitably drift, and the diversity of detection requirements will cause the sonar detection viewpoint to change, and then produce sonar images with nonlinear intensity differences. These images often do not have explicit gradient information, and the gray level tends to be equalized. It is difficult to use artificially designed matching algorithms and is not conducive to the generalization of the model, unless the area has obvious landmark features. Considering the huge difference between the acoustic imaging mechanism and the optical imaging mechanism, applying the optical matching algorithm developed around the image intensity to the acoustic image will also lose a certain degree of robustness and accuracy. At present, a dynamic research idea is to use deep convolutional neural network technology to perform sonar image matching, but the current research is reflected in the stage of regional similarity evaluation, and there is no method to detect and describe features, the universal mode of matching and filtering completes the sonar image matching task of nonlinear intensity difference.

In order to solve this problem, this paper proposes a feature matching algorithm for sonar images with nonlinear intensity differences. There are three main contributions in this paper.

Firstly, phase consistency (PC) is used to detect feature points instead of gray level and gradient value, and the number and repeatability of feature points are considered.

Secondly, the similarity evaluation of deep convolutional neural network output is applied to feature description. The network is constructed by 2-channel network and has better performance than the traditional Siamese network. This method not only greatly improves the stability of feature detection, but also overcomes the limitations of sonar image gray and gradient information in feature description.

Thirdly, we use the classical and state-of-the-art matching approaches for the comparison to analyze their performance in sonar image matching tasks, involving manually designed, deep learning based and transformer-based approaches.

III. DETAILED METHODOLOGY

The method is mainly described from three stages: feature detection, feature description and feature matching.

*A. Feature detection by phase information*

Morrone and Owens believe that in the image, the feature can be perceived at the point of the maximum phase of the Fourier component, the Fourier component is in phase at the step point of the square wave and the peak and trough of the triangular wave, and this property tends to be stable in scale [14]. On this basis, they proposed the concept of PC. In the case of one dimension, the phase consistency of a certain position is expressed as follows:

$$PC(x) = \max_{\bar{\phi}(x) \in [0, 2\pi]} \frac{\sum_n A_n \cos(\phi_n(x) - \bar{\phi}(x))}{\sum_n A_n} \quad (1)$$

where $A_n(x)$ represents the amplitude of the *nth* Fourier component, $\phi_n(x)$ represents the phase angle of the *nth* Fourier component, so that $\bar{\phi}(x)$ maximizing the equation is the amplitude weighted average local phase angle of all Fourier terms at the point under consideration.

The value of the PC can be used to measure the significance of the feature. 1 means that the feature is very significant, and 0 means that the feature is not significant. However, phase consistency is usually difficult to calculate. In [14], the author proposed to find the peak value in the local energy function to equivalently calculate the point of maximum phase consistency. This is the default local energy function and phase consistency function. So theoretically, the peak in the local energy will correspond to the peak in the phase consistency. Li [15] applied the PC principle to the multimodal image matching task sensitive to intensity and gradient, and developed a robust RIFT algorithm. The feature detection in this paper is based on the detector of RIFT. The calculation of PC is as follows:

$$f(t) = \sum_{n=1}^{\infty} A_n \sin(2\pi n f_0 t + \varphi_0) = \sum_{n=1}^{\infty} A_n \sin(\varphi_n(t)) \quad (2)$$

$$pc(t) = \max_{\varnothing} \frac{\sum_{n=1}^{\infty} A_n \sin(\varphi_n(t) - \varnothing)}{\sum_{n=1}^{\infty} A_n} \quad (3)$$

where $A$ represents the amplitude, $f$ represents the frequency, and $N$ represents the number of sin signals.

*B. Deep convolution feature similarity estimation*

The idea of using Siamese network for image matching and similarity evaluation was systematically proposed in [16]. The branch of the network can be regarded as a descriptor calculation module, and the core of the top network is the similarity evaluation. The basic structural framework is shown in Fig. 1.

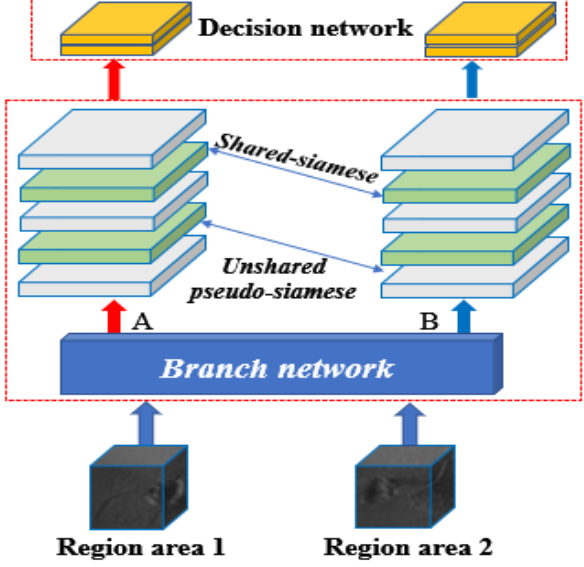

**Fig. 1.** The architecture of basic Siamese network.

To determine the region area 1 and region area 2 in the sonar image, firstly need to construct a network mapping function $G_W(X)$, and then use region area 1 and area 2 as the parameter independent variables $X_1$, $X_2$, we can get $G_W(X_1)$, $G_W(X_2)$, and that is, the feature vector used to evaluate whether $X_1$ and $X_2$ are similar is obtained. Next, construct the *Loss* as follows:

$$E_W(X_1, X_2) = \|G_W(X_1) - G_W(X_2)\| \qquad (4)$$

Different from the basic Siamese network, there is no direct descriptor concept in the architecture of the 2-channel network, and the training architecture is improved. Just treat the two input blocks as one 2-channel image and feed it directly to the first convolutional layer of the network.

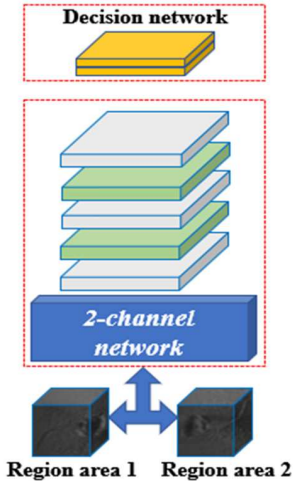

**Fig. 2.** The architecture of 2-channel network.

In this case, the bottom of the network consists of a series of convolution layers, relu layers, and maximum pooling layers. The output of the bottom part is then used as the input of the top module, which contains only a fully connected linear decision, which makes a layer have one output. Therefore, compared with the basic network architecture, the 2-channel network is a lightweight structure with greater flexibility and faster training speed, and is more suitable for underwater engineering technology research. The parameters and details of the 2-channel network are described in [16].

*C. Matching process*

We map the global and local similarity evaluation results established by the 2-channel network model to the local feature points extracted by the PC detector to determine whether the image area matches. The input of the model is a pair of images of the same area detected from different viewpoints. In order to improve the real-time performance of the algorithm, in the image preprocessing stage, the two image blocks are combined into a 2-channel image.

We strictly align the sonar region areas through an expert system composed of locating data and acoustic echo data, and then encode and segment the images according to the alignment method, so that we could construct data samples in real time and statistically. Combine the slices into positive samples on the alignment sequence, and then throw them out of order. Now these two randomly combined slices are statistically mismatched, that is, negative samples, the construction idea is described in Fig. 3. Next, we divide the training set, test set and validation set according to a certain proportion. It is worth mentioning that in the process of constructing data samples, we did not set the size of each input patch, but freely combined them with different sizes, such as 16x16, 32x32, and 64x64, to obtain better model performance. In addition, data enhancement operation and the strategy for variable learning rate are introduced to reduce the impact of overfitting.

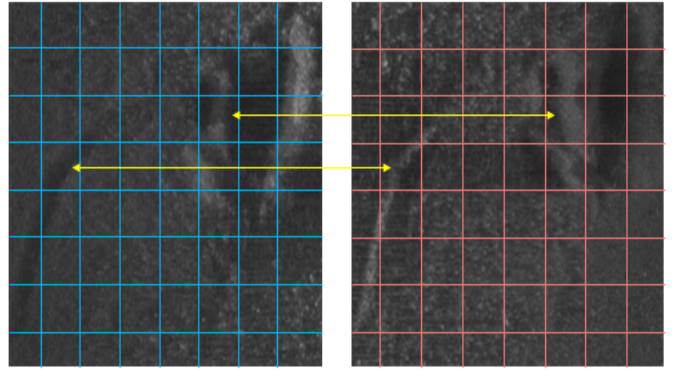

**Fig. 3.** Schematic diagram of training sample construction.

The matching method is to separately intercept the region area of the pre-sized area centered on the feature keypoints detected according to the phase information in Figure A and Figure B, and then input them into network model to determine whether they match or not .If the output of the network model is 1, then it is judged that the two region areas are matched, that is, the matching problem is equivalent to the binary classification problem.

The procedure of the nonlinear intensity sonar image matching algorithm is expressed as follows:

---

**ALGORITHM 1**: Nonlinear intensity sonar image matching algorithm

---

**Input**: Nonlinear intensity sonar image pair *A* and *B*
**Output**: Matching result of image *A* and image *B*
**Process**:
 1. Strictly align the sonar image pair and divide the region areas
 2. Slice region images to construct labeled samples for training
 3. Extract deep convolution features and make similarity evaluation
 4. Map evaluation results to the feature keypoints detected by PC
 5. Match keypoints and eliminate the wrong matches
**END**

---

## IV. EXPERIMENTAL SETTING

### A. Experiment Data

We selected a group of SSS images with nonlinear intensity caused by the heading of AUV turning 180 degrees from north to south. The detailed tracking path is depicted in Fig. 4.

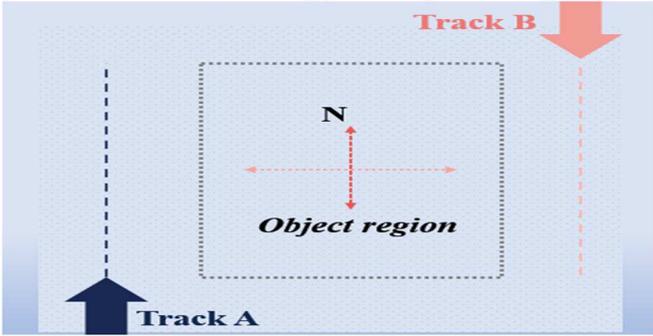

**Fig. 4.** Repeated detection of the object region in the north-south track.

The SSS images on the Fig. 5 is obtained by Deep Vision AB company [17] using the DeepEye 680D in Lake Vättern, Sweden. We select a group of nonlinear intensity sonar image regions after strict alignment for the subsequent matching test, and the amplified intensity difference is shown in Fig. 6.

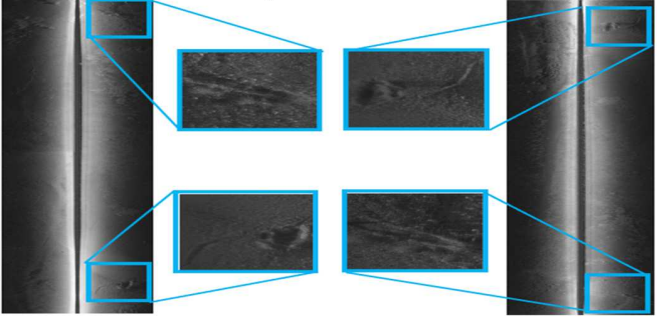

**Fig. 5.** SSS image pairs with nonlinear intensity difference.

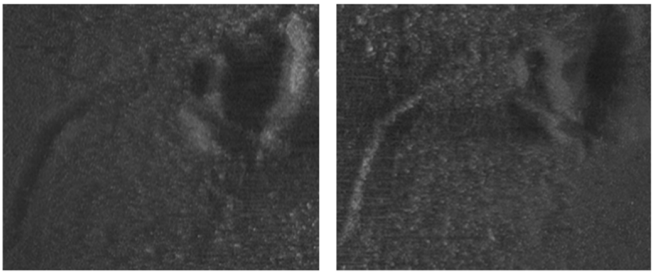

**Fig. 6.** The architecture of basic Siamese network.

### B. Comparative approaches

In the subsequent experiments, we introduced image matching approaches SIFT, ORB, BRISK [18], SuperPoint [19] and LoFTR [20] for comparison. These methods are classic and advanced methods in matching tasks and have shown good performance in many scene matching tasks.

1) **SIFT**
   This is a scale invariant feature detection and extraction algorithm, in which the descriptor is obtained by considering the pixels in the field of key point location radius. It is one of the most classical algorithms in image matching.
2) **ORB**
   This algorithm is built on the FAST keypoint detector and it is less affected by image noise. Its advantage is time performance, and it is widely used in industry.
3) **BRISK**
   It has good rotation invariance, scale invariance and good robustness, especially when applied to large blurred images, it performs very well, and it has obvious advantages in matching speed. The BRISK algorithm mainly uses FAST9-16 for feature point detection, and multi-scale expression by constructing an image pyramid.
4) **SuperPoint**
   This is a new type of deep learning based matching algorithm in the matching field. It proposes a self-supervised framework to train key points and feature descriptors suitable for multi-view geometric problems.
   Its experimental test shows that a simple and efficient CNN can solve the problem of detection and description of sparse key points, and the entire system can well complete the task of optical vision matching.
5) **LoFTR**
   It is a newly developed matching method based on Transformer technology, showing great advantages in matching accuracy and real-time. It proposes to build pixel-level dense matches at the coarse level and refine the good matches at the fine level. Its performance is better than most of the current matching algorithms.

### C. Testing environment

All methods were implemented under the Windows 10 operating system using Python 3.7 with an Intel Core i7-9700 3.00GHz processor, 16GB of physical memory, and one NVIDIA GeForce RTX2070s graphics card. SIFT, ORB and BRISK are implemented based on openCV-Python tools [21]. In order to maximize the matching performance of the above methods, we have adopted their original parameter settings, in which the matching distance threshold ($d_{ratio}$) of SIFT, ORB and BRISK is set to 0.85 and the matchinng mode is K-Nearest Neighbor (KNN).

## V. EXPERIMENTAL RESULTS

We compared the overall matching effect of classical and advanced matching methods on sonar images with nonlinear intensity differences, and further proved the adaptability of the detector based on phase information. Finally, we tried our method on the original image without strict alignment.

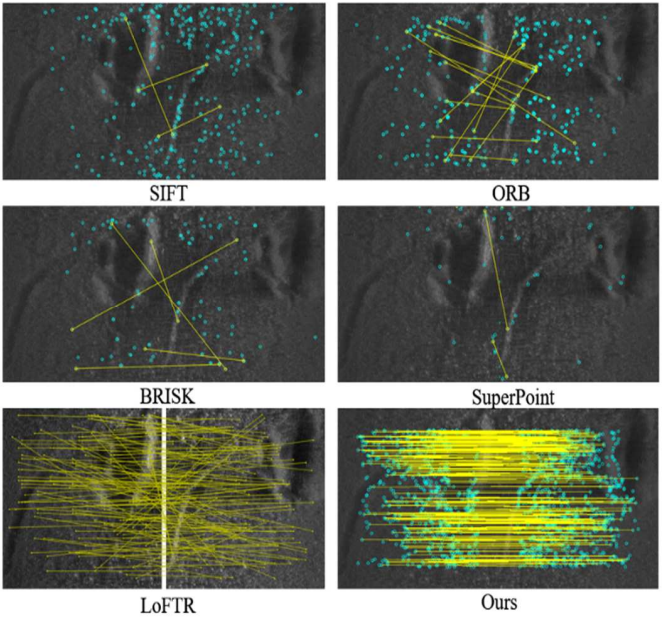

**Fig. 7.** The schematic diagram of overall matching effects comparison.

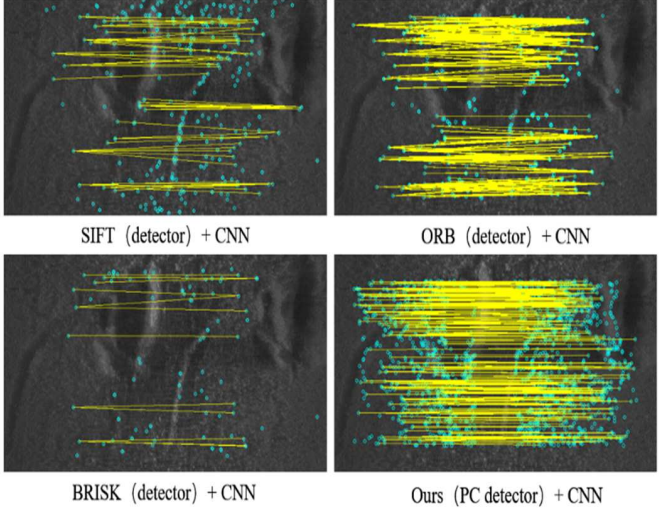

**Fig. 8.** Matching effects of CNN descriptors combined with various detectors.

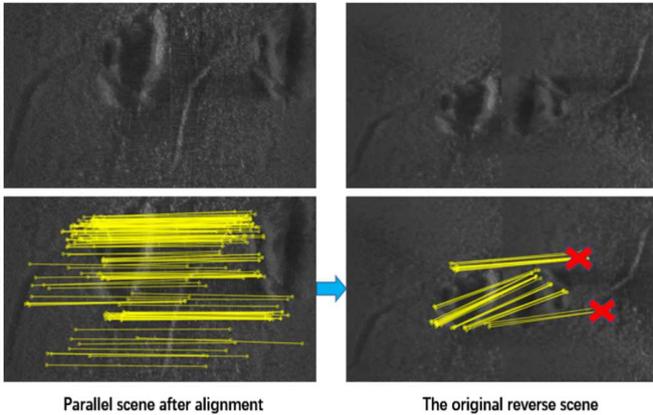

**Fig. 9.** The effects of our method directly applied to the original reverse scene, the above are the original images, and the below are the matching effects of our method.

It could be seen from Fig. 7, for sonar image pairs with large nonlinear intensity differences, only our proposed approach could robustly complete the matching process. Other methods contribute little, and the feature points we detected are also relatively concentratedly distributed around the target area, with good aggregation, reducing the time to traverse the overall situation. Additionally, we also compared the detection effects of three hand-designed classic detectors and the detector based on phase information, as illustrated in Fig. 8. It could be seen that when the key points detected by the three classic detectors are combined with the CNN similarity descriptor, the matching effects will have crossover errors, while the phase information-based detector could obtain more accurate and robust results. In the end, we directly tried the matching effect of our model in the original reverse scene, and found that the number of matching pairs was very small, and there were mismatches, as displayed in Fig. 9. Since other methods cannot obtain effective matching results, this article does not discuss the comparison of timeliness.

In the future, we will further improve the generalization ability of our method to adapt to large-angle changing scenarios. We are no longer limited to the standard north-south tracking, but use a more flexible heading angle for detection. In addition, we plan to introduce the idea of transfer learning and collect more sonar images to relieve the pressure of sample shortage, while capturing some other types of sonar images, such as FLS images, to verify our method.

## VI. CONCLUSION

Aiming at the sonar image matching problem with nonlinear intensity difference, a combined matching method is proposed that uses the phase information to detect feature points, and then uses the similarity of deep convolution features to describe the feature points. Our method correlates the similarity evaluation of sonar images with the coordinates of local feature points, and the results show that when the image coordinate points and ping are used to assist in aligning the images, the method can better deal with the nonlinear intensity between the sonar image pairs difference issues. Our method does not make any assumptions about the topography, bottom quality and other characteristics, but its angle selection is not flexible enough at present. In the future, we will continue to optimize the detector and similarity evaluation model, and combine them for different engineering application scenarios to achieve the purpose of matching complex sonar images fastly.